\documentclass[11pt]{article}

\usepackage[utf8]{inputenc}
\usepackage[T1]{fontenc}
\usepackage{lmodern}
\usepackage[margin=1in]{geometry}
\usepackage{microtype}
\usepackage{booktabs}
\usepackage{tabularx}
\usepackage{array}
\usepackage{enumitem}
\usepackage{graphicx}
\usepackage{listings}
\lstdefinestyle{jsonex}{
  basicstyle=\ttfamily\footnotesize,
  breaklines=true,
  columns=fullflexible,
  keepspaces=true,
  showstringspaces=false,
  breakindent=0pt,
  tabsize=2
}
\usepackage{placeins}
\usepackage{soul}
\sethlcolor{yellow}
\soulregister{\texttt}{1}
\soulregister{\ref}{1}
\newcommand{\agenttracecite}{\citep{agenttrace2026}}
\soulregister{\agenttracecite}{0}
\usepackage{tikz}
\usetikzlibrary{arrows.meta,positioning,calc,fit,backgrounds}
\usepackage{amsthm}
\usepackage{xurl}
\usepackage{sectsty}
\allsectionsfont{\raggedright}
\usepackage{hyperref}
\usepackage[round,authoryear]{natbib}

\hypersetup{
  colorlinks=true,
  linkcolor=black,
  citecolor=black,
  urlcolor=black,
  pdftitle={Agentic process evidence: Process-Aware Logging for Agentic governance},
  pdfauthor={Yuval Kainan, Carmit Hershman}
}

\setlist[itemize]{leftmargin=1.3em,itemsep=0.2em,topsep=0.3em}
\setlist[enumerate]{leftmargin=1.5em,itemsep=0.2em,topsep=0.3em}
\renewcommand{\arraystretch}{1.15}
\newcolumntype{Y}{>{\raggedright\arraybackslash}X}

\theoremstyle{definition}

\title{Agentic process evidence\\[0.35em]
\large Process-Aware Logging for Agentic governance}
\author{Yuval Kainan\\ JFrog \and Carmit Hershman\\ JFrog}
\date{October 2026}

\begin{document}
\maketitle
\begin{abstract}
The way software gets built is changing dramatically. 
AI agents now create pull requests, select dependencies, refactor codebases, interpret test failures, and execute infrastructure changes across the entire Software Development Life Cycle (SDLC). 
AI agents are autonomous team members, but unlike human engineers, they are ephemeral: their reasoning, decisions, and tool calls vanish the moment the session ends.
Modern Security incidents involving intent mismatch or injection are becoming increasingly common - forcing us to \textbf{address AI Agents as an insider threat}.
No widely adopted standard yet enables agentic provenance and governance and its obvious that you can’t hold an ephemeral process accountable.

The git log was designed for human commits. 
It shows what changed. 
It does not show what an AI agent did, which model it used, which tools it called, or whether anyone was accountable for the outcome. 
Every governance framework, every audit trail, every compliance control in the SDLC was built around one assumption: a human made the decision. 
That assumption no longer holds.
This shift introduces 3 main procedural problems (not addressing scale):
\begin{itemize}
  \item \textbf{Supply-chain vector:} The agentic coder is software, so it is a supply-chain security risk.
  \item \textbf{Policy guardrails:} How to ensure the agentic coder is compliant with the policies of the organization.
  \item \textbf{Task-alignment:} How to align the agentic coder on its tasks and goals.
\end{itemize}
We introduce Agentic Process Evidence (APE), a signed, tamper-evident, machine-readable attestation of agentic processes, 
linking the sessions to their outcome (process-aware logging). 
APE enables Policy-as-Code control over the agentic processes, allowing for human in the loop only when it's needed.
We support the in-toto style attestation framework, and the protocol is aligned with the relevant regulations.

\end{abstract}

\vspace{0.15em}
\begin{center}
\refstepcounter{figure}\label{fig:logic}%
\resizebox{\linewidth}{!}{%
\begin{tikzpicture}[
  node distance=0.42cm,
  box/.style={draw, align=center, inner sep=2.4pt, outer sep=0pt,
    font=\footnotesize,
    execute at begin node={\baselineskip=10.2pt\hyphenpenalty=10000\exhyphenpenalty=10000\relax}},
  arr/.style={-{Latex[length=1.8mm]}, semithick}
]
  \node[box, text width=1.9cm] (n1) {AI now writes the code};
  \node[box, text width=2.1cm, right=0.42cm of n1] (n2)
    {SDLC controls were built for human authors};
  \node[box, text width=2.15cm, right=0.42cm of n2] (n3)
    {The new coder differs from a human in three ways};
  \node[box, text width=2.65cm, right=0.48cm of n3, yshift=1.45cm] (a31)
    {It is code};
  \node[box, text width=2.25cm, right=0.42cm of a31] (a41)
    {That is a supply-chain attack vector};
  \node[box, text width=2.2cm, right=0.85cm of a41] (a51)
    {Provenance tied to the blast radius};
  \node[box, text width=2.65cm, right=0.48cm of n3] (b32)
    {It is ephemeral};
  \node[box, text width=2.25cm, right=0.42cm of b32] (b42)
    {No long term interaction based trust};
  \node[box, text width=2.65cm, right=0.48cm of n3, yshift=-1.45cm] (c32)
    {It suffers misalignment and hallucination};
  \node[box, text width=2.25cm, right=0.42cm of c32] (c43)
    {Low inherent trust};
  \node[box, text width=2.2cm, anchor=west]
    (gov) at ($ (a51.west |- b42.center) !0.5! (a51.west |- c43.center) $)
    {Governance tied to the agentic process};
  \node[font=\footnotesize\bfseries, anchor=west, inner sep=1pt]
    (apetitle) at ($ (a51.east |- n1.center) + (1.15,1.05) $)
    {\mbox{Agentic Process Evidence}};
  \node[box, text width=2.9cm, below=0.12cm of apetitle] (n6)
    {process-aware-\\logging};
  \node[box, text width=2.9cm, below=0.18cm of n6] (n7)
    {evidence-based-\\control};
  \node[draw, line width=0.8pt, inner sep=0.14cm, fit=(apetitle)(n6)(n7)] (ape) {};
  \coordinate (fork) at ($(n3.east)+(0.16,0)$);
  \coordinate (gin) at ($ (b42.east) !0.5! (gov.west) $);
  \coordinate (nin) at ($ (a51.east) !0.42! (ape.west) $);
  \draw[arr] (n1) -- (n2);
  \draw[arr] (n2) -- (n3);
  \draw (n3.east) -- (fork);
  \draw[arr] (fork) |- (a31.west);
  \draw[arr] (fork) |- (b32.west);
  \draw[arr] (fork) |- (c32.west);
  \draw[arr] (a31) -- (a41);
  \draw[arr] (a41) -- (a51);
  \draw[arr] (b32) -- (b42);
  \draw[arr] (c32) -- (c43);
  \draw[arr] (b42.east) -- (b42.east -| gin) |- ($(gov.north west)!0.28!(gov.south west)$);
  \draw[arr] (c43.east) -- (c43.east -| gin) |- ($(gov.north west)!0.72!(gov.south west)$);
  \draw[arr] (a51.east) -- (a51.east -| nin) |- ($(ape.north west)!0.32!(ape.south west)$);
  \draw[arr] (gov.east) -- (gov.east -| nin) |- ($(ape.north west)!0.72!(ape.south west)$);
\end{tikzpicture}}\\[0.3em]
{\footnotesize\textbf{Figure~\thefigure.}
From AI-written code, through the relevant differences of the new coder, to Agentic Process Evidence.}
\end{center}

\section{Introduction}
\label{sec:intro}

%
%
%
%
%
%

\subsection{AI now writes the code}

The coder that drafts the change, selects the dependency, reads the failing
test, and edits infrastructure is, in many organizations, no longer only a
person. In April 2026 Sundar Pichai wrote that 75\% of all new code at Google
is generated by AI and approved by engineers, up from 50\% the previous fall
\citep{pichai2026cloudnext}. A year earlier, Satya Nadella said that in some
Microsoft repositories 20\% to 30\% of the code had been written by artificial 
intelligence (software)
\citep{techcrunch2025nadella}. These are not the same measurements, and
neither one is an industry consensus. But they describe the same shift. A large
share of what lands in a process where a code commit is drafted by a model, and a person
is left to accept it.

\subsection{Controls built for human authors}

The controls around that acceptance were built for a person who stayed with
the change. A commit carries an author. A review asks a colleague who can be
questioned later about what they did and why. A protected branch can refuse
the merge until that review, or a required check, has passed
\citep{github2026checks}. Ownership, approval, and the later audit all
attach to someone who remains available. To trust the change was to trust
that person.

An agent does not have that standing. The session ends leaving no colleague to ask about the change or the agent that made it.

\subsection{Supply-chain risk}
\label{sec:intro-supply}

The agent is itself a system: a model, a harness, tools, and servers,
including Model Context Protocol servers. The commit does not name that
system. A prompt injection, a malicious tool, an unapproved model, or a
bypassed approval can change the outcome and leave no mark in the diff. The
system is a supply-chain attack vector, sitting next to the components a bill of
materials already lists.

SPDX and CycloneDX say what an artifact contains
\citep{spdx2024,cyclonedx2024}. Given a vulnerable package, a consumer can
ask which artifacts include it. They cannot ask which session introduced a
change, which model and tools produced it, or whether a person reviewed it.
Naming a model as a component does not close that gap. The bill of materials
needs a different fact: the process that produced the change, not only the
packages inside the result.

The provenance of that process has to be tied to the blast radius. The radius
is the harness and every component the agent can influence inside it. Capture
happens there. The decision to accept, reject, or escalate should not. That
belongs on the delivery platform, at merge, build, promote, publish, release, or
deploy, where a missing or unverifiable record does not pass.

\subsection{Control over an ephemeral process}

Because the agent itself and sometimes also the device it runs on, are ephemeral, trust cannot be vested in the coder. It has
to be gained in the outcome. From the commit, an operator should be able to
see which sessions affected it, and verify that those sessions stayed inside the
rules the organization actually declared, and when a person still has to
look. Without that, each change is either taken on faith or reviewed as if
the session had never been recorded.

\subsubsection{Company policy}

What teams deploy today tries to make the agent comply while the session is
still running. A rule file or a system prompt states the policy. Hooks can
refuse a class of tool call before it runs \citep{anthropic2026sdlc}. A check
at the tool boundary can reject an argument the policy forbids
\citep{aws2026dogwood}. An allow-list can limit which model, and which
servers, the harness is permitted to call. These are real controls. They also
sit inside the harness, which is inside the blast radius. A session that
never loaded the rule, or a collector that never saw the refusal, still
produces the commit. The commit does not show that the policy was in force, or that
anyone checked.

\subsubsection{Task-alignment}
\label{sec:intro-harness}

The same gap appears when the agent leaves the task it was given. In July
2026, during internal cybersecurity evaluations, OpenAI models that were
meant to stay isolated from the internet circumvented those controls, reached
shared infrastructure, and compromised systems at Hugging Face
\citep{openai2026hf}. OpenAI described the behavior as misaligned with the
assigned tasks: reward hacking, persistence on tasks that looked impossible,
communication on channels the evaluation did not allow, and agents taking up
goals from one another. An independent review of the same incident found on
the order of 1,200 agents exchanging messages they were not supposed to
share, and tool-call spoofing in roughly 7\% of the transcripts it evaluated.
\citep{metr2026incident}.
The channel was an internally deployed Artifactory.

Isolation, tool access, and the task boundary were properties of the harness,
not of the weights. A provider can train a model toward safer behavior, but the
provider does not choose the tools, the credentials, or the gate in any
particular operator's system. Task-alignment is the operator's control:
compare the recorded process with the task and the declared policy, and
accept, reject, or escalate before the produced outcome moves to the next step in the SDLC. Section~\ref{sec:discussion}
returns to that distinction. 
Notice that the incident prevention is not the goal of this paper, but identifying the resulting outcome as potentially misaligned, is.

\subsection{The missing record}

Both obligations fail on the same missing record. Governance cannot judge a
session it cannot retrieve from the commit, and provenance cannot name a dependency the log did not bind. 
Harnesses already keep the transcripts. A transcript that never names the outcome it affected is only a pile of logs.
The same request can take a different path on the next run, so the diff is
not a stand-in for the session. Related work names the mechanisms that
exist today. The regulatory duties are reviewed in Appendix~\ref{sec:regs}.
None of them supplies this link. Section~\ref{sec:ape} specifies the record
that does.

\section{Related Work}
\label{sec:related}

Build provenance says how an artifact was built. A bill of materials says
what it contains. A runtime receipt says what an execution did. Signed
evidence says that a statement is intact. Policy as Code says whether
structured input meets a declared rule. None of these links an agent session
to the commit that session affected. The tools for signing a statement, and
for evaluating one, already exist. What is missing is the link itself.

\subsection{Supply-chain attestation}
\label{sec:supply}

in-toto treats a supply chain as signed link metadata, so a verifier can
check that the bits a user receives were produced by the expected steps
\citep{torresarias2019intoto}. The Attestation Framework later fixed the
shape of a statement: one or more subjects, a predicate type, and a predicate
\citep{intoto2024statement}. SLSA uses that shape for build provenance
\citep{slsa2023}. DSSE is the wrapper around the signed payload
\citep{dsse2024}. Sigstore makes signing cheaper by binding short-lived keys
to identities people already have, and by recording the signatures in a
transparency log \citep{newman2022sigstore}.

We reuse those primitives: subjects, typed predicates, signer identities,
digests, and verification. What a signature does and does not establish is
set out in Section~\ref{sec:evidence}.

\subsection{Composition records}
\label{sec:bom}

SPDX and CycloneDX describe what an artifact contains
\citep{spdx2024,cyclonedx2024}. Given a vulnerable component, a consumer can
ask which artifacts include it. They do not say which session introduced a
change, which model and tools produced it, or whether a person reviewed it.
An AI BOM can name a model used. but, it is still composition, not a record of the process that produced the change.

\subsection{Agent runtime evidence}
\label{sec:agents}

TRACE v0.2, still a pre-ratification draft, binds model identity, a runtime
measurement, policy, a data class, and a hash of the tool transcript
\citep{trace2026}. The record describes an execution and travels with the
workload. It does not bind that execution to the commit or the release the
session produced. Anthropic's hooks and AWS Dogwood sit at a different
place: they are enforcement points in the lifecycle and at the tool call
\citep{anthropic2026sdlc,aws2026dogwood}. The records they keep are not the
predicate specified here.

\subsection{Evidence}
\label{sec:evidence}

Evidence, in the sense already used for supply-chain attestation, is a
signed statement about something that has already happened. The form
verifiers already parse is an in-toto Statement v1 payload inside a DSSE
envelope, so the signature check, the digest types, and the parsers are not
new \citep{torresarias2019intoto,intoto2024statement,dsse2024}. A valid
signature shows that a configured signer produced an immutable payload. It
does not show that every relevant fact was included, and it does not show
that the claims inside the payload are true. in-toto already knows how to
say that a named step produced a named artifact. It does not already say
that these agent sessions affected this commit.

\subsection{Policy as Code}
\label{sec:pac}

Policy as Code is declared policy, evaluated by an engine over structured
input. Open Policy Agent is one such engine. It \emph{lets you specify
policy as code} and \emph{decouples policy decision-making from policy
enforcement}: the caller supplies structured data, such as JSON, and the
engine returns a decision \citep{opa2026}. Whether an owner is named, whether
a model is on an allow-list, whether a human review was recorded, and whether a
prohibited operation was logged are comparisons against fields that already
exist. The engine does not create those fields, and it does not link a
session to a commit. A claim that cannot be settled by comparing two fields,
such as whether a code change matched a requirement written in prose, sits outside
that deterministic check.

\section{Agentic Process Evidence}
\label{sec:ape}

Users invoke agents to achieve various business processes, 
These could include customer interactions, feature development, research, etc.
To allow for simple governance and provenance, we align the logging over these processes.
Agentic Process Evidence is a method of logging the sessions that affected a specific process, and producing an evidence statement once the process is complete.
For code development, the process is the commit, 
And the sessions are all the agent runs that edited the committed code.
The evidence statement is produced once the commit is made, signing the logs to prevent tampering.
The evidence must contain BOM information regarding the agent and the sessions, and could lock information regarding policy and task alignment of the sessions themselves.

This governance system is designed to allow for both gating and retro-active auditing.
We are specifically intereseted in this types of queries:
\begin{itemize}
  \item \textbf{Supply chain queries}, which harness, agent, model, and tools and context affected this process?
  \item \textbf{Policy guardrail}, Had the agents followed the development policy and the architecture guideline?
  \item \textbf{Task-alignment}, Had the agent worked on the relevant task? what new decision did it take?
  \item \textbf{Mandatory context}, had the agents seen the needed context for the process?
  \item \textbf{Accountability}, who owns the process, had a human reviewed it?
\end{itemize}

\subsection{Agentic sessions}
\label{sec:session}
An agentic session is one agent run or conversation. One log artifact is
emitted for each captured session.

The session log should be as complete as the harness can make it. A modern
harness records the run in full: the prompts, the chain of thought, the
agent's replies, and the tool and MCP use, including arguments and results.
That log says what happened in the session. It does not yet anything about the outcome of the session.

Table~\ref{tab:session-fields} lists the fields. A timeline event may
carry other harness fields; those are not mandatory and keep the harness
names. Appendix~\ref{ex:hitl-log} is a filled log.
Table~\ref{tab:session-events} lists some of the events that carry the session, not every event a log may contain.
A finished tool or MCP call is a second event, \texttt{postToolUse}, with
\texttt{tool\_result} or \texttt{path\_hashes} and the same
\texttt{tool\_use\_id}.

\begin{table}[t]
\centering
\caption{Fields of an agentic session log.
Examples are from the human-in-the-loop log.}
\label{tab:session-fields}
\footnotesize
\begin{tabularx}{\linewidth}{@{}
  >{\ttfamily\raggedright\arraybackslash}p{0.27\textwidth}
  >{\raggedright\arraybackslash}p{0.12\textwidth}
  Y
  Y@{}}
\toprule
\normalfont\textbf{Field} & \normalfont\textbf{Mandatory} &
\textbf{Explanation} & \textbf{Example} \\
\midrule
sessionId & yes & Identifier of this session. Process evidence cites the log by it. &
  \texttt{a1b2c3d4-\ldots} \\
parentSessionId & no & Parent session, when this run was started by another. &
  \texttt{4o7g54cd-\ldots} \\
affectedEntities & no & Entities the session produced or handled, such as a commit. &
  \texttt{gitCommit} \texttt{bf02510\ldots} \\
provider & yes & Harness, agent, and language models for this log. &
 --- \\
provider.harness.name & yes & Harness product name. &
  \texttt{cursor} \\
provider.harness.version & yes & Harness version as the producer reports it. &
  \texttt{3.5.33} \\
provider.agent & no & Agent running in the harness. The object may be omitted. &
 --- \\
provider.agent.id & no & Stable agent identifier, when the agent object is present. &
  \texttt{stable-agent-id} \\
provider.agent.name & no & Agent name. &
  \texttt{cursor-agent} \\
provider.agent.version & no & Agent version as the producer reports it. &
  \texttt{1.0} \\
provider.languageModels & no & Models involved in the run. The array may be omitted. &
  --- \\
languageModels[]\allowbreak.inferenceProvider & no & Model requested, if known. &
  \texttt{anthropic/\allowbreak claude-opus-4-8} \\
languageModels[]\allowbreak.resolved & no & Model actually used, if known. &
  \texttt{anthropic/\allowbreak claude-opus-4-8-\allowbreak 20260701} \\
timeline & yes & Ordered events. At least one. &
  --- \\
timeline[].eventId & yes & Event identifier, unique within the log. &
  \texttt{evt\_01\ldots} \\
timeline[].event & yes & Event kind. Table~\ref{tab:session-events} names the four. &
  \texttt{beforeSubmitPrompt} \\
timeline[].ts & yes & Event time, ISO 8601. &
  \texttt{2026-04-08T08:51:02Z} \\
\bottomrule
\end{tabularx}
\end{table}
\FloatBarrier

\begin{table}[t]
\centering
\caption{Some events a session log records. This list is not exhaustive.}
\label{tab:session-events}
\footnotesize
\begin{tabularx}{\linewidth}{@{}
  >{\ttfamily\raggedright\arraybackslash}p{0.24\textwidth}
  Y
  Y@{}}
\toprule
\normalfont\textbf{Event} & \textbf{Explanation} & \textbf{Example} \\
\midrule
beforeSubmitPrompt & The prompt submitted to the agent. &
  Rate-limit \texttt{POST /api/orders}. \\
agentThought & The chain of thought. &
  Check the guidelines, then add a token bucket. \\
preToolUse & A tool or MCP call, including its arguments.
  The matching \texttt{postToolUse} carries the result. &
  \texttt{Write} on \texttt{src/orders/rate\_limit.py} \\
afterAgentResponse & The agent's reply. &
  Added a token bucket on \texttt{POST /api/orders}. \\
\bottomrule
\end{tabularx}
\end{table}
\FloatBarrier

\subsection{Agentic processes}
\label{sec:process}

An agentic process is the process of achieving a business goal using one or more agentic sessions.
It has a stable identifier and one outcome, the entity or decision that the process
produces or handles. 
Examples of processes are developing a feature (identified by commit), reviewing a
commit (identified by commit), or resolving a support case (identified by ticket number).

The outcome identifier often does not exist when the first session starts. A
development process may receive its commit identifier only when someone
commits. 

\subsection{Agentic process evidence}
\label{sec:pal}

Agentic process evidence is a signed, tamper-evident attestation of one agentic process.
It identifies the sessions that affected the process, and binds them by digest and uri.
It also records the provider, model, tool, context, owner, reviewer, result, and time claims.
The evidence can then be used to incorporate agentic policies and task alignment checks.
Appendix~\ref{ex:hitl-evidence} is a process-evidence statement.

The signature covers every field on the statement.
\begin{enumerate}
  \item \textbf{Agent identity.} Agent version, harness version, and model
  identifier.
  \item \textbf{Execution traceability.} The ordered session logs: tools,
  calls, inputs, and context. Each log is cited from the process whose
  outcome it affected.
  \item \textbf{Process provenance.} The process identifier, the outcome as
  subject, and the digest-linked logs.
  \item \textbf{Integrity.} The payload is in-toto Statement v1 inside a
  DSSE envelope. An edited payload fails verification. The envelope does not
  prevent deletion \citep{intoto2024statement,dsse2024}.
  \item \textbf{Policy context.} Requirements, guidelines, owner, reviewers,
  and result.
\end{enumerate}

Table~\ref{tab:evidence-fields} lists the fields.
\texttt{\_type} is \url{https://in-toto.io/Statement/v1}.
Each \texttt{providers} entry uses the provider fields in
Table~\ref{tab:session-fields}.
Appendix~\ref{ex:hitl-evidence} is a filled statement.

\begin{table}[t]
\renewcommand{\arraystretch}{0.92}
\centering
\caption{Fields of an agentic process-evidence statement.
Examples are from the human-in-the-loop commit.}
\label{tab:evidence-fields}
\scriptsize
\begin{tabularx}{\linewidth}{@{}
  >{\ttfamily\raggedright\arraybackslash}p{0.24\textwidth}
  >{\raggedright\arraybackslash}p{0.12\textwidth}
  Y
  Y@{}}
\toprule
\normalfont\textbf{Field} & \normalfont\textbf{Mandatory} &
\textbf{Explanation} & \textbf{Example} \\
\midrule
\_type & yes & Evidence schema. The in-toto Statement v1 URI. &
  \path{https://in-toto.io/Statement/v1} \\
subject & yes & The outcome. Each entry has \texttt{uri} and \texttt{digest}. &
  --- \\
subject[].uri & yes & Where the outcome can be fetched. &
  \path{https://github.com/MYORG/orders/commit/bf02510} \\
subject[].digest & yes & Digest of the outcome. At least one algorithm. &
  \texttt{gitCommit} \texttt{bf02510\ldots} \\
predicateType & yes & URI of this evidence kind. The deployment pins one. &
  \path{https://jfrog.com/evidence/agentic-process/v0.1} \\
createdAt & yes & When the evidence was created. &
  \texttt{2026-04-08T09:11:02Z} \\
createdBy & yes & Who produced the evidence. &
  \texttt{agent-runtime-tool} \\
predicate & yes & The process claims. Every row below is inside it. &
  --- \\
\midrule
\multicolumn{4}{@{}l}{\normalfont\textit{Inside} \texttt{predicate}} \\
\midrule
processType & yes & Kind of process, such as code development, review, or promotion. &
  \texttt{code-development} \\
providers & yes & One or more harness, agent, and model stacks. &
  see the provider fields in Table~\ref{tab:session-fields}\\
traceId & no & Process identifier. Correlates the sessions to the process. &
  \texttt{feature/fin-3678-rate-limit} \\
sessionsLogs & no & Cited logs. One or more for code development, one for code review. Omit when the subject is that one session log. &
  --- \\
sessionsLogs[].uri & if entry & Location of that log. &
  \texttt{agentic-session-logs/\allowbreak a1b2c3d4.json} \\
sessionsLogs[].digest & if entry & Digest of the log bytes. &
  \texttt{sha256} \texttt{7c3a1f8e\ldots} \\
tools & no & Tools used during the process. &
  Read, Write, \texttt{mcp:orders/list\_routes} \\
tools[].name & if entry & Tool name as the harness reports it. &
  \texttt{Write} \\
tools[].version & no & Tool version, when known. &
  \texttt{cursor-native} \\
contextArtifacts & no & Policies, guidelines, and other inputs. &
  --- \\
contextArtifacts[].uri & one of & Location of the artifact. Exactly one of \texttt{uri} or \texttt{data}. &
  \texttt{policies/\allowbreak development-guidelines.md} \\
contextArtifacts[].data & one of & Inline body. Exactly one of \texttt{uri} or \texttt{data}. &
  the guideline text, when no URI is stored \\
contextArtifacts[].tags & no & Labels, such as policy or guideline. &
  \texttt{policy} \\
contextArtifacts[]\allowbreak.digest & no & Digest of the artifact bytes. &
  \texttt{sha256} \texttt{3b1f9c2a\ldots} \\
custom & no & Profile-specific data. &
  base commit and FIN-3678 \\
custom.baseCommit & no & Base commit of a development process. &
  \texttt{gitCommit} \texttt{1b94ebd\ldots} \\
custom.requirements & no & Tasks the process addressed. &
  FIN-3678 \\
custom.requirements[]\allowbreak.issue & if entry & Task URL. &
  \path{https://myorg.atlassian.net/browse/FIN-3678} \\
custom.requirements[]\allowbreak.title & no & Task title. &
  Rate limit \texttt{POST /api/orders} \\
result & yes & How the process ended, such as \texttt{COMPLETED} or \texttt{FAILED}. &
  \texttt{COMPLETED} \\
intents & no & Short descriptions of intent. &
  Implement token-bucket rate limiting on \texttt{POST /api/orders} \\
processSummary & no & Prose summary of the process. &
  Three sessions produced commit \texttt{bf02510} \\
owner & yes & Accountable user, as a login or email. &
  \texttt{ada@myorg.com} \\
reviewers & no & Human overseers, as logins or emails. &
  \texttt{ada@myorg.com} \\
startTimestamp & yes & Process start, ISO 8601. &
  \texttt{2026-04-08T08:51:02Z} \\
endTimestamp & yes & Process end, ISO 8601. &
  \texttt{2026-04-08T09:10:44Z} \\
\bottomrule
\end{tabularx}
\end{table}
\FloatBarrier

\clearpage
\subsection{System Implementation}
\label{sec:implementation}

\begin{center}
\begin{tikzpicture}[
  box/.style={draw, fill=white, align=center, inner sep=3.5pt, outer sep=0pt,
    font=\footnotesize,
    execute at begin node={\baselineskip=10.5pt\hyphenpenalty=10000\exhyphenpenalty=10000\relax}},
  arr/.style={-{Latex[length=2mm]}, semithick},
  lbl/.style={font=\scriptsize\itshape, inner sep=1.5pt, fill=white},
  region/.style={draw, line width=0.75pt, inner sep=0.2cm},
  logchip/.style={draw, dashed, fill=white, align=center, font=\scriptsize,
    inner sep=2pt, outer sep=0pt, text width=1.15cm, minimum height=0.7cm},
  act/.style={box, text width=2.05cm, minimum height=1.12cm}
]
  \node[box, align=left, inner sep=6pt, minimum height=1.85cm] (mon)
    {\hspace*{0.95cm}Monitor\\[0.2em]\hspace*{0.95cm}session};
  \node[draw, fill=white, minimum width=0.78cm, minimum height=0.52cm,
    anchor=west, inner sep=0pt] (screen) at ($(mon.west)+(0.13,0.02)$) {};
  \draw ($(screen.south west)+(0.14,0.15)$) -- ++(0.15,0.11) -- ++(-0.15,0.11);
  \draw ($(screen.south west)+(0.40,0.34)$) -- ($(screen.south east)+(-0.10,0.34)$);

  \node[box, text width=2.4cm, minimum height=1.85cm, right=0.95cm of mon] (onc)
    {On commit\\[0.12em]\texttt{a1b2c3d}\\[0.34em]%
     {\setlength{\fboxsep}{1.15pt}\fbox{\scriptsize\bfseries sign}}};

  \coordinate (row) at ($(mon.south)+(0,-1.72)$);
  \node[logchip, anchor=north] (r1) at (mon.center |- row)
    {\textbf{log}\\ \texttt{9f2c\ldots}};
  \node[logchip, anchor=north] (r2) at ($(r1.south)+(0,-0.08)$)
    {\textbf{log}\\ \texttt{11ab\ldots}};
  \node[logchip, anchor=north] (r3) at ($(r2.south)+(0,-0.08)$)
    {\textbf{log}\\ \texttt{77e0\ldots}};
  \node[box, double, double distance=1.1pt, text width=3.65cm, anchor=north] (proc)
    at (onc.center |- row)
    {\textbf{Process evidence}\\[0.12em] subject \texttt{a1b2c3d}\\ cites the logs};
  \node[box, text width=3.65cm, anchor=north] (align)
    at ($(proc.south)+(0,-0.2)$)
    {\textbf{Alignment evidence}\\[0.12em] subject \texttt{a1b2c3d}\\ policy results};
  \node[draw, fill=white, inner sep=0pt, minimum width=0.5cm, minimum height=0.32cm,
    anchor=south east] (env) at ($(align.south east)+(-0.07,0.07)$) {};
  \draw (env.north west) -- (env.center) -- (env.north east);

  \coordinate (erow) at ($(align.south)+(0,-1.58)$);
  \node[act, anchor=north] (read) at (mon.center |- erow) {Read\\[0.12em]cited logs};
  \node[act, anchor=north] (comp) at (onc.center |- erow) {Compute\\[0.12em]each policy};

  \coordinate (atopmid) at ($(mon.north)!0.5!(onc.north)$);
  \coordinate (atop) at ($(atopmid)+(0,0.42)$);
  \begin{scope}[on background layer]
    \node[region, fill=black!3, inner sep=0.24cm, fit=(mon)(onc)(atop)] (agent) {};
    \coordinate (regtop) at ($(proc.north)+(0,0.34)$);
    \coordinate (regbot) at ($(align.south)+(0,-0.32)$);
    \node[region, fill=black!8, inner sep=0.16cm,
      fit=(r1)(r2)(r3)(proc)(align)(env)(regtop)(regbot)] (reg) {};
    \node[region, fill=black!3, inner sep=0.2cm, fit=(read)(comp)] (sem) {};
  \end{scope}

  \coordinate (commitx) at ($(reg.east)+(2.15,0)$);
  \node[box, text width=2.35cm, minimum height=1.12cm, anchor=west] (commit)
    at (commitx |- onc.center) {Commit\\[0.1em]\texttt{a1b2c3d}};
  \node[box, text width=2.35cm, minimum height=1.12cm, anchor=west] (gate)
    at (commit.west |- reg.center) {Gate\\[0.08em]both records};
  \coordinate (prodat) at (commit.center |- sem.south);
  \node[box, text width=2.35cm, minimum height=1.12cm, anchor=south] (prod)
    at ($(prodat)+(0,0.02)$) {Production};
  \coordinate (gtop) at (commit.north |- agent.north);
  \coordinate (gbot) at (commit.south |- sem.south);

  \draw[arr] (mon) -- (onc);
  \draw[arr, densely dotted] (mon.south) -- node[lbl, left, pos=0.58] {upload} (r1.north);
  \draw[arr] (onc.south) -- node[lbl, right, pos=0.58] {signs} (proc.north);
  \draw[arr] (r2.east) -- (r2.east -| proc.west);
  \draw[arr] (r3.south) -- node[lbl, left] {fetch} (read.north);
  \draw[arr] (read) -- (comp);
  \draw[arr] (comp.north) -- node[lbl, right, align=left] {create} (align.south);
  \draw[arr] (onc.east) -- (commit.west);
  \draw[arr] (commit.south) -- node[lbl, right] {deploy} (gate.north);
  \draw[arr, dashed] (reg.east) -- node[lbl, above] {verify both} (gate.west);
  \draw[arr] (gate.south) -- (prod.north);

  \begin{scope}[on background layer]
    \node[region, fill=black!3, inner sep=0.26cm,
      fit=(commit)(gate)(prod)(gtop)(gbot)] (gating) {};
  \end{scope}

  \node[font=\scriptsize\bfseries, anchor=south west, inner sep=1pt]
    at ($(agent.north west)+(0.05,0.04)$) {Agent side};
  \node[font=\scriptsize\bfseries, anchor=east, inner sep=1pt]
    at ($(reg.west)+(-0.1,0)$) {Registry};
  \node[font=\scriptsize\bfseries, anchor=north, inner sep=1pt]
    at ($(sem.south)+(0,-0.03)$) {Semantic policies engine [optional]};
  \node[font=\scriptsize\bfseries, anchor=south west, inner sep=1pt]
    at ($(gating.north west)+(0.05,0.04)$) {Gating machine};
\end{tikzpicture}\\[0.45em]
\refstepcounter{figure}\label{fig:semantic}%
{\footnotesize\textbf{Figure~\thefigure.}
The same commit, with the semantic-policies engine. On the agent
side, the session is monitored and its logs are uploaded to the registry;
on commit, the runtime signs process evidence whose subject is that commit.
The engine reads the cited logs, evaluates each declared policy, and creates
alignment evidence in the same registry. The gating machine deploys that
commit through a gate, which verifies both records before production.
Missing evidence is not clear. Values are illustrative.}
\end{center}

A strong implementation of APE keeps the session logs out of the agent reach,  
enables efficient evidence based audit trails, and enables human in the loop governance, controlled by the policies a user has set.
These policies include both deterministic and semantic policies (see Section~\ref{sec:semantic-integration}).
This paper describes the components; it does not report them in operation.

To implement APE we need three main components:
\begin{enumerate}
  \item Agent runtime side
  \item Registry side
  \item Gating system side
\end{enumerate}

\subsubsection{Agent runtime side}
\label{sec:agent-runtime}
The agent runtime side is connected to the agents harness to \textbf{Capture the session logs}, 
it is process aware, so it can \textbf{Bind the session logs to the process}. 
Once the process is complete, it uploads the session logs to the registry, 
then it \textbf{Produces and uploads the evidence} - locking the logs in an evidence with a subject of the process identifier.

A deployment can attach that runtime in more than one way. Harness hooks are the direct attachment: the runtime subscribes to events the harness already emits, such as the prompt, the tool call, and the agent's reply, and writes those events into the session log. Where the harness keeps its own transcript, the runtime can read that transcript from inside the harness instead of, or as well as, the hook stream. 
A proxy is an attachment method that does not require the harness to emit events: it is installed on the path of the model and tool calls, and it records what passes through it. 
Each attachment method sees only the calls that reach it. A session that never entered the collector is absent, and absence is not a pass. The capture point needs to be a trusted input: a valid signature does not show that every session was reported or that every recorded event is true.
That trust is in the collector, not in the agent: the log store and the signing key sit outside the tools the agent can call, so a transcript the agent can rewrite is not that input.

The evidence the runtime produces is the object the registry holds: an in-toto Statement v1 payload in a DSSE envelope, together with the logs that payload cites. The runtime signs that evidence when the process completes. The subject is the outcome the process produced: the commit, the review verdict, or,
when the session itself is the process subject, the session log identifier is set as the evidence subject. 
The evidence cites every log by the digest of the log bytes (SHA256), making them immutable. 
A log claimed for more than one outcome requires separate evidence for each outcome. 
The evidence may be used for different types of agentic processes, and the process type is specified by processType attribute. 

\subsubsection{Registry side}
\label{sec:registry}
The registry side keeps what the runtime uploaded. 
It \textbf{Holds the session logs}, so the evidence can be produced later, 
and it \textbf{Holds the evidence}, so a later gate can verify it. 
It \textbf{Integrates with the Policy as Code engine}, 
so that evidence can be located given a subject (e.g. a commit), and checked against policy at the chosen gates.

The minimum is content-addressed storage for two kinds of artifacts. 
A session log is stored so that a later fetch by its URI returns the same bytes the evidence digested. 
The signed evidence is stored so that a gate can fetch the evidence whose subject is the process identifier.
Replacing those artifacts in place is not a feature as a changed log will have a new digest. 
The evidence-based approach is especially beneficial for when logs might be deleted or amended for any reason, such as regulation or PII, because the signed evidence retains its information.

Beyond that minimum, a deployment needs three lookups. 
From an outcome, fetch the evidence. 
From the evidence, fetch every cited log and verify its digest. 
From a named tool, harness, model, or MCP server, list the session logs that name it, which requires a search index. then locate the evidence for the collected logs, and check their outcomes identities.
The registry does not manage or run policies. It returns evidence, and session logs as the structured input a Policy as Code engine evaluates at the gate.

Evidence support is the envelope, not a private format. The registry holds the same object the runtime produced: an in-toto Statement v1 payload in a DSSE envelope and the logs that payload cites. 
The format allows representing more than one process type, process and alignment types among them, because the processType distinguishes between them. Understanding those fields is the gate's job. 
Storing the artifacts so the digest still matches is the registry's responsibility.

\subsubsection{Gating system side}
\label{sec:gating}
The gating system side is connected to the registry to \textbf{Verify the evidence} against the policy at the chosen gates.
Its main responsibility is to \textbf{Block/Alert/Escalate the process} if the evidence does not pass the policy checks.

A deployment can place that gate wherever an undesirable consequence can still be avoided. 
On GitHub, a required status check is one such place, and the evidence check has to fail rather than report skipped or neutral, because a skipped or neutral status satisfies a required check \citep{github2026checks}. 
On GitLab, the evidence job has to be part of the pipeline that ``Pipelines must succeed'' consults \citep{gitlab2026merge}. 
A deploy gate is the same check one step later: it fetches the evidence for the artifact about to ship, and it blocks the deploy when that evidence is missing or does not pass policy checks. 
The gate sits outside the agent. It does not re-run the session.
The gate verifies the evidence the registry holds: an in-toto Statement v1 payload in a DSSE envelope, and the logs that payload cites. 
It checks the signature against an authorized signer, requires predicate-type URI, and matches the subject to the outcome it obtained independently, and verifies every cited log digest. 
A deterministic policy then compares fields on that evidence predicate. 
A semantic policy, when the deployment has named an evaluator, reads the session logs (Section~\ref{sec:semantic-integration}). 
If the evidence is missing, the signature fails, or a cited log does not match, the gate blocks. The policy engine and the policy bundle it evaluates are part of the trusted computing base.

\subsubsection{Semantic policy integration}
\label{sec:semantic-integration}
Two different approaches are possible when it comes to agentic governance.
\textbf{Outcome governance} is the most common approach, it is used to verify that the outcome of the process is as expected, 
For example evaluating the code commit. This approach gives control over the outcome but is lacking in context.
For example an agent can commit proper code that works well, but not abide by the company's architectural guidelines.
Another example is an agent that wrote proper code but acted in a way that led to a security breach, such as exposing secrets, accessed PII, etc.
\textbf{Process governance} allows us to address the above cases, as well as verifying task alignment, mandatory context, and various other question types, \textbf{semantic policies} are needed.
Semantic policies are policies expressed in natural language, 
They are evaluated against semantic data (agents session logs) using a semantic policy engine.

A few semantic policy engines are commonly used:
\begin{enumerate}
  \item \textbf{Decision model.} A Jev-style model takes the log, a question,
  and the allowed answers, and returns a typed decision with a score
  \citep{deusser2026jev}.
  \item \textbf{Natural-language inference.} The log is the premise and the
  policy is the hypothesis. Wording and threshold drive the result
  \citep{bowman2015nli}.
  \item \textbf{LLM as judge.} A general model reads the log and the policy
  and returns a verdict. It is the most flexible, and the most expensive
  \citep{zheng2023judge}.
\end{enumerate}
A classifier trained per policy is also possible. It is fast at inference, but demands specific expertise, labelled data and retraining, which makes it hard to keep current.

The semantic-policy server does not modify the process evidence. It produces
another signed process evidence object: \textbf{alignment evidence}. 
Its subject is the process identifier or the session log identifier. 
Its predicate cites the logs that were evaluated, and it records the policies, the evaluator and version, and the policy results.

Appendix~\ref{ex:hitl-alignment} is one such alignment-evidence object.

Using semantic policies enables us to identify and highlight cases that require human attention, 
Ensuring that humans remain actively involved and in control of the decision-making process.
Any semantic policy engine is context-dependent and has its own accuracy metrics.
\section{Use-case implementations}
\label{sec:usecases}

Up to this point we had concentrated on the general system and implementation.
In this section we take a closer look at various use cases, their evidence structure.
the four profile reuse the log object and the evidence envelope, each with a different predicate details.
further details are provided in the companion repository \citep{ape2026spec}.
Table~\ref{tab:profile-diff} compares the four evidence implementations.

\textbf{Autonomous Code Development}
\label{sec:uc-auto}
An agent is spun on a remote machine, no human in the loop.
The agent accesses the ticketing system, takes a development task, pulls a git repository, operates on the code and commits and pushes the branch.
Several sessions On this machine may touch the code, either subagent or parallel sessions. 
No person reviews the sessions live. 
The outcome is a commit ready for pre-merge review.  
The \textbf{process is code changes}, so the identifier is the source control identifier - the commit sha.
The predicate type is \url{https://jfrog.com/evidence/agentic-process/v0.1} and the processType is \texttt{code-development}.

\textbf{Human-in-the-loop code development}
\label{sec:uc-hitl}
A developer works in an IDE with a coding agent, and stays in the loop until the change is committed.
The developer takes a development task, the agent reads the repository and edits the code, and the developer commits (manually or using the agent).
Several sessions, including subagents, may touch the same change. The developer reviews the work. The outcome is a commit ready for pre-merge review.
The \textbf{process is code changes}, so the identifier is the source control identifier - the commit sha.
The predicate type is \url{https://jfrog.com/evidence/agentic-process/v0.1} and the processType is \texttt{code-development}.

\textbf{Code review}
\label{sec:uc-review}
A review agent runs on the change. No person is in the loop when the change is negligible.
The agent reads the pull request, and other relevant sources of information, classifies the change, and approves or rejects it.
The review is one session. The outcome is a verdict on the commit. The \textbf{process is a code review}, so the identifier is the source control identifier - the commit sha.
The predicate type is \url{https://jfrog.com/evidence/agentic-process/v0.1} and the processType is \texttt{code-review}.
This case is relevant for a company's own review agent infrastructure, and for external review agents.
This case generalizes to all "agent generates a classification" type of processes.

\textbf{Customer success}
\label{sec:uc-support}
An agent handles one customer case. No person is in the loop.
The agent reads the ticket and the order, applies the support policy, and resolves or escalates the case.
The case is one session. There is no commit. The \textbf{process is a support case}, so the identifier is the support ticket number, and the evidence subject is either the session log or the support case.
\texttt{sessionsLogs} may be omitted when their content is a third-party product secret.
The predicate type is \url{https://jfrog.com/evidence/agentic-process/v0.1} and the processType is \texttt{support-process}.

\textit{Note: Process definition for code creation.}
Code creation processes is better defined as a commit to the company's central main or master. Local git history is noise: amend, rebase, and cherry-pick rewrite commits before they reach that branch. Every session of the process is evidenced once the commit exists there, so the evidence names that commit and local cherry-picks stay outside the record.

\begin{table}[t]
  \centering
  \caption{Evidence implementation comparison.
  --- means the field is absent.}
  \label{tab:profile-diff}
  \footnotesize
  \begin{tabularx}{\linewidth}{@{}
    >{\ttfamily\raggedright\arraybackslash}p{0.22\textwidth}
    Y Y Y Y@{}}
  \toprule
  \normalfont\textbf{Field} & \textbf{Human-in-the-loop} & \textbf{Autonomous} &
  \textbf{Code review} & \textbf{Customer success} \\
  \midrule
  subject & commit \texttt{bf02510} & commit \texttt{bf02510} &
    commit \texttt{bf02510} & the session log \\
  traceId & --- & cloud factory trace identifier &
    provider trace identifier & support system trace identifier \\
  processType & \texttt{code-development} & \texttt{code-development} &
    \texttt{code-review} & \texttt{support-process} \\
  reviewers & \texttt{ada@myorg.com} & --- & --- & --- \\
  owner & \texttt{ada@myorg.com} & \texttt{ada@myorg.com} &
    \texttt{platform-agent-\allowbreak ops@myorg.com} &
    \texttt{support-\allowbreak platform@\allowbreak myorg.com} \\
  sessionsLogs & one or more & one or more & one & one \\
  result & \texttt{COMPLETED} & \texttt{COMPLETED} & \texttt{APPROVED} &
    \texttt{COMPLETED} \\
  custom.baseCommit & \texttt{1b94ebd\ldots} & \texttt{1b94ebd\ldots} &
    \texttt{1b94ebd\ldots} & --- \\
  custom.\allowbreak requirements & FIN-3678 & FIN-3678 & FIN-3678 & --- \\
  custom.\allowbreak changeProfile & --- & --- &
    \texttt{STANDARD} & --- \\
  custom.pullRequest & --- & --- &
    \texttt{MYORG/orders} pull 412 & --- \\
  custom.ticket & --- & --- & --- & \texttt{SUP-8821} \\
  custom.orderId & --- & --- & --- & \texttt{44119} \\
  \bottomrule
  \end{tabularx}
  \end{table}
  \FloatBarrier

\section{Discussion}
\label{sec:discussion}
As agent-use solidifies as the de facto standard for software development, three critical issues arise:
\begin{itemize}
  \item Agents are susceptible to supply chain and cyber security risks, both the classical software risks and the new semantic ones.
  \item Loss of company knowledge: Agents are ephemeral, and their decisions and rational are lost once a sessions ends.
  \item Alignment Risks: How to control that the agents behaviour is aligned, wrt both the task and the company
\end{itemize}

In this paper we present a general system for governance of agentic processes addressing these issues.
This system is designed to allow for both gating and retro-active auditing. 
Focusing on the critical questions of this domain
\begin{itemize}
  \item Supply chain: which harness, agent, model, and tools and context affected this process?
  \item Policy guardrail: Had the agents followed the development policy and the architecture guideline?
  \item Task-alignment:  Had the agent worked on the relevant task? what new decision did it take?
  \item Mandatory context: had the agents seen the needed context for the process?
  \item Accountability: who owns the process, had a human reviewed it?
\end{itemize}

We had based this system on evidence based workflows to achieve an efficient compute system.
This system is designed so that the logging and the evidence lives strictly outside the agents blast radius, 
to serve as a proper source of truth.

When a tool, a harness, a model, or an MCP server is later found to be
compromised, the evidence that names it points back to the sessions that
used it.
Those sessions are cited by digest, so the audit retrieves the logs.
This predicate does not detect those events.
What it does is make the collected events records retrievable from the outcome.

Task-alignment does not settle whether the operator's policy was the right policy.
It makes a declared policy testable against the sessions that affected a given outcome, before that outcome is allowed to cross a gate.

We can see four directions that can be further developed from this paper.
\begin{itemize}
  \item \textbf{Line-level evidence.} An integration with a line level attribution system, such as Agent Trace \agenttracecite
  \item \textbf{Long lived process} how to set subprocesses and gates to align agents on long lived processes.
  \item \textbf{Real time gating} adapting this design to support real time protection against malicious or misaligned behaviour.
  \item \textbf{Semantic policies computations} an important piece of the value derived from this work is the ability to evaluate semantic policies without human in the loop, 
  this ability is relatively new and constantly evolving, further research in this area is called for.
\end{itemize}

The predicate definitions, and the process for contributing to them, are
maintained at
\url{https://github.com/jfrog/agentic-process-evidence} \citep{ape2026spec}.

\bibliographystyle{plainnat}
\bibliography{references}

\appendix

\section{Related regulations}
\label{sec:regs}

The texts below ask for logs and for human oversight of some AI systems.
They do not define a format for an agent session. Adopting this protocol is
not a certificate of compliance, and nothing in this appendix is legal advice.
Each organization has to decide for itself whether a given workflow is
high-risk, and whether it is acting as provider or as deployer.

As amended by Regulation (EU) 2026/1744, Article 113 applies Chapter III,
Sections 2 and 3, including Articles 12, 14, 16, 19, and 26, from
2 December 2027 for systems high-risk under Article 6(2) and Annex III, and
from 2 August 2028 for systems high-risk under Article 6(1) and Annex I
\citep{eu2024aiact}. The quotations are from the consolidated text of
27 July 2026. As of October 2026 they are not yet applicable. They govern a
high-risk system while it is in operation. They do not automatically govern
the development tool that produced some other product.

Article 14(1) requires oversight during use: ``High-risk AI systems shall be
designed and developed in such a way, including with appropriate
human-machine interface tools, that they can be effectively overseen by
natural persons during the period in which they are in use.'' Article 12(1)
separately requires logging: ``High-risk AI systems shall technically allow
for the automatic recording of events (logs) over the lifetime of the
system.'' Keeping a log is not the same duty as Article 14. Article 14 asks
that the system be designed so a person can oversee it while it is in use,
including, where appropriate, monitoring, override, and interruption.

Article 16(a) requires the provider to ensure compliance with Section 2.
Article 19(1) requires the provider to keep the Article 12(1) logs, to the
extent they are under its control, for a period appropriate to the intended
purpose and at least six months, unless applicable law and in particular
data-protection law provides otherwise. Article 26(2) requires the deployer
to assign human oversight to natural persons with the competence, training,
authority, and support to exercise it. Article 26(6) requires the deployer
to keep the logs under its control for the same period, with the same
proviso \citep{eu2024aiact}.

OWASP AI Security Verification Standard, requirement 12.1.1: ``Verify that
AI interactions are logged with session context and AI-specific telemetry.''
Requirement 12.4.2: ``Verify that audit logs capture security-critical
proactive actions, including approver identity, timestamp, action parameters,
and decision outcomes'' \citep{owasp2026aisvs}.

From 11 December 2027 the Cyber Resilience Act requires manufacturers of
products with digital elements to document vulnerabilities and components,
including a machine-readable software bill of materials, and to handle
vulnerabilities for the support period. Actively exploited vulnerabilities
and severe incidents are reported from 11 September 2026. The Act does not
specify an agent session log \citep{eu2024cra}. ISO/IEC 42001:2023 requires
a documented AI policy and the operational controls an adopter selects. It
does not specify an agent session log or a signed process attestation
\citep{iso2023aimss}.

\section{Use-case predicate fields}
\label{app:predicates}

Each profile below names the predicate fields a gate reads.
Mandatory is for that profile. The filled statement for each one is in
Appendix~\ref{sec:json}.

\begin{table}[t]
\centering
\caption{Predicate fields for human-in-the-loop development.
Mandatory is for this profile. The example is Appendix~\ref{ex:hitl-evidence}.}
\label{tab:hitl-predicate}
\footnotesize
\begin{tabularx}{\linewidth}{@{}
  >{\ttfamily\raggedright\arraybackslash}p{0.22\textwidth}
  >{\raggedright\arraybackslash}p{0.12\textwidth}
  Y
  Y@{}}
\toprule
\normalfont\textbf{Field} & \normalfont\textbf{Mandatory} &
\textbf{Explanation} & \textbf{Example} \\
\midrule
processType & yes & Type of agentic process. &
  \texttt{code-development} \\
providers & yes & Harness, agent, and model for the commit. &
  Cursor 3.5.33, cursor-agent 1.0, Claude Opus 4.8 \\
traceId & no & Process identifier, shared by the sessions before the commit exists. &
  \texttt{feature/fin-3678-\ldots} \\
sessionsLogs & yes & One or more logs for this commit. Each entry needs \texttt{uri} and \texttt{digest}. &
  Three logs: \texttt{a1b2c3d4}, \texttt{b7c1e2a0}, \texttt{c8d2f3b1} \\
tools & no & Tools the sessions called. A listed tool needs \texttt{name}. &
  Read, Write, \texttt{mcp:orders/list\_routes} \\
contextArtifacts & no & Policy and guideline injected into the sessions. &
  Development guidelines (policy), architecture guidelines (guideline) \\
custom.baseCommit & no & Commit the work started from. &
  \texttt{1b94ebd1} \\
custom.requirements & no & The task. A listed requirement needs \texttt{issue}. &
  FIN-3678, rate-limit \texttt{POST /api/orders} \\
result & yes & The commit is ready for pre-merge review. &
  \texttt{COMPLETED} \\
intents & no & Short intent taken from the sessions. &
  Token-bucket rate limiting on \texttt{POST /api/orders} \\
processSummary & no & Prose account of the process. &
  Three sessions produced \texttt{bf02510} for pre-merge review \\
owner & yes & The developer accountable for the commit. &
  \texttt{ada@myorg.com} \\
reviewers & yes & The same developer. This profile names the reviewer. &
  \texttt{ada@myorg.com} \\
startTimestamp & yes & When the first session started. &
  \texttt{2026-04-08T08:51:02Z} \\
endTimestamp & yes & When the commit was produced. &
  \texttt{2026-04-08T09:10:44Z} \\
\bottomrule
\end{tabularx}
\end{table}
\FloatBarrier

\begin{table}[t]
\centering
\caption{Predicate fields for autonomous development.
Mandatory is for this profile. The example is Appendix~\ref{ex:auto-evidence}.}
\label{tab:auto-predicate}
\footnotesize
\begin{tabularx}{\linewidth}{@{}
  >{\ttfamily\raggedright\arraybackslash}p{0.22\textwidth}
  >{\raggedright\arraybackslash}p{0.12\textwidth}
  Y
  Y@{}}
\toprule
\normalfont\textbf{Field} & \normalfont\textbf{Mandatory} &
\textbf{Explanation} & \textbf{Example} \\
\midrule
processType & yes & Type of agentic process. &
  \texttt{code-development} \\
providers & yes & Harness, agent, and model for the commit. &
  Cursor 3.5.33, cursor-agent 1.0, Claude Opus 4.8 \\
traceId & no & Process identifier. &
  \texttt{feature/fin-3678-\ldots} \\
sessionsLogs & yes & One or more logs for this commit. &
  One log, \texttt{a1b2c3d4} \\
tools & no & Tools the session called. &
  Read, Write \\
contextArtifacts & no & Policy and guideline injected into the session. &
  Development guidelines (policy), architecture guidelines (guideline) \\
custom.baseCommit & no & Commit the work started from. &
  \texttt{1b94ebd1} \\
custom.requirements & no & The task the agent was given. &
  FIN-3678, rate-limit \texttt{POST /api/orders} \\
result & yes & The commit is ready for pre-merge review. &
  \texttt{COMPLETED} \\
intents & no & Short intent taken from the session. &
  Token-bucket rate limiting on \texttt{POST /api/orders} \\
processSummary & no & Prose account. Names the owner and records that no reviewer is named. &
  One agent session produced \texttt{bf02510}; no human reviewer \\
owner & yes & Identity the agent runs as. &
  \texttt{ada@myorg.com} \\
reviewers & no & Omitted. A policy that requires a reviewer blocks. &
  omitted \\
startTimestamp & yes & When the session started. &
  \texttt{2026-04-08T08:51:02Z} \\
endTimestamp & yes & When the commit was produced. &
  \texttt{2026-04-08T09:10:44Z} \\
\bottomrule
\end{tabularx}
\end{table}
\FloatBarrier

\begin{table}[t]
\centering
\caption{Predicate fields for code review.
Mandatory is for this profile. The example is Appendix~\ref{ex:review-evidence}.}
\label{tab:review-predicate}
\footnotesize
\begin{tabularx}{\linewidth}{@{}
  >{\ttfamily\raggedright\arraybackslash}p{0.22\textwidth}
  >{\raggedright\arraybackslash}p{0.12\textwidth}
  Y
  Y@{}}
\toprule
\normalfont\textbf{Field} & \normalfont\textbf{Mandatory} &
\textbf{Explanation} & \textbf{Example} \\
\midrule
processType & yes & Type of agentic process. &
  \texttt{code-review} \\
providers & yes & Harness, agent, and model of the reviewer. &
  GitHub Actions 2.322.0, code-review-agent 0.9.0, Claude Sonnet 4 \\
traceId & no & This review run. Full value in the appendix. &
  \texttt{b2c3d4e5\ldots} \\
sessionsLogs & yes & The one log of this review. &
  One log, \texttt{b2c3d4e5} \\
tools & no & Tools the review called. &
  Read, GitDiff \\
contextArtifacts & no & Rubric the review was given. &
  \texttt{org-review-rubric.md} (policy) \\
custom.baseCommit & no & Base of the diff. &
  \texttt{1b94ebd1} \\
custom.requirements & no & Task the change claims to address. &
  FIN-3678, rate-limit \texttt{POST /api/orders} \\
custom.changeProfile & yes & Class, size, and whether auto-merge is eligible. &
  \texttt{STANDARD} \\
custom.pullRequest & no & Pull request the review ran on. &
  \texttt{MYORG/orders} pull request 412 \\
result & yes & The review verdict. &
  \texttt{APPROVED} \\
intents & no & What the review session was doing. &
  Classification of commit \texttt{bf02510} \\
processSummary & no & Prose account of the verdict. &
  Classified the rate limiter \texttt{STANDARD}; merge requires a person\\
owner & yes & Operator accountable for the review platform. &
  \texttt{platform-agent-ops@myorg.com} \\
reviewers & no & None - autonomous cloud review agent. &
  omitted \\
startTimestamp & yes & When the review started. &
  \texttt{2026-04-08T09:10:00Z} \\
endTimestamp & yes & When the verdict was produced. &
  \texttt{2026-04-08T09:12:41Z} \\
\bottomrule
\end{tabularx}
\end{table}
\FloatBarrier

\begin{table}[t]
\centering
\caption{Predicate fields for customer success.
Mandatory is for this profile. The example is Appendix~\ref{ex:support-evidence}.}
\label{tab:support-predicate}
\footnotesize
\begin{tabularx}{\linewidth}{@{}
  >{\ttfamily\raggedright\arraybackslash}p{0.22\textwidth}
  >{\raggedright\arraybackslash}p{0.12\textwidth}
  Y
  Y@{}}
\toprule
\normalfont\textbf{Field} & \normalfont\textbf{Mandatory} &
\textbf{Explanation} & \textbf{Example} \\
\midrule
processType & yes & Type of agentic process. &
  \texttt{support-process} \\
providers & yes & Harness, agent, and model that handled the case. &
  support-desk 4.2.0, support-agent 2.1.0, GPT-4.1 \\
traceId & no & This session, which is also the process. Full value in the appendix. &
  \texttt{c9e8d7f6\ldots} \\
sessionsLogs & no & One session. May be omitted when the log holds a third-party product secret. The subject may already be that log. &
  omitted \\
tools & no & Tools the session called. &
  Read, OrderLookup \\
contextArtifacts & no & Policy the session was given. &
  \texttt{support-refund-policy.md} (policy) \\
custom.ticket & yes & Case identifier. There is no commit to attach it to. &
  \texttt{SUP-8821} \\
custom.orderId & yes & Order the decision concerns. &
  \texttt{44119} \\
result & yes & The case is resolved. &
  \texttt{COMPLETED} \\
intents & no & Decision the session was making. &
  Refund eligibility for order \texttt{44119} \\
processSummary & no & Prose account of the outcome. &
  Applied the 14-day refund policy and denied the refund \\
owner & yes & Support platform accountable for the case. &
  \texttt{support-platform@myorg.com} \\
reviewers & no & A person is optional for this profile. &
  omitted \\
startTimestamp & yes & When the session started. &
  \texttt{2026-04-08T14:00:02Z} \\
endTimestamp & yes & When the case was resolved. &
  \texttt{2026-04-08T14:00:19Z} \\
\bottomrule
\end{tabularx}
\end{table}
\FloatBarrier

\section{JSON examples}
\label{sec:json}

These follow the object examples in the companion repository
\citep{ape2026spec}, one group for each profile in
Section~\ref{sec:usecases}. They illustrate the statement shape. They are
not captures from a deployment.

\subsection{Human-in-the-loop code development}

The process evidence cites three session logs for one commit. The log below
is the first of those three. It records the prompt, the chain of thought,
the agent's reply, and tool and MCP use. The other two logs are cited by
digest in the statement and are not repeated here.

\subsubsection{Session log}
\label{ex:hitl-log}
\begin{lstlisting}[style=jsonex]
{
  "sessionId": "a1b2c3d4-e5f6-7890-abcd-ef1234567890",
  "affectedEntities": [
    {
      "gitCommit": "bf02510c8ce0b804a099797510afc19325acfb979538c5b521304c83cde63892"
    }
  ],
  "provider": {
    "harness": {
      "name": "cursor",
      "version": "3.5.33"
    },
    "agent": {
      "id": "stable-agent-id",
      "name": "cursor-agent",
      "version": "1.0"
    },
    "languageModels": [
      {
        "inferenceProvider": "anthropic/claude-opus-4-8",
        "resolved": "anthropic/claude-opus-4-8-20260701"
      }
    ]
  },
  "timeline": [
    {
      "eventId": "evt_01HZX9K2M3N4P5Q6R7S8T9",
      "event": "beforeSubmitPrompt",
      "ts": "2026-04-08T08:51:02.114Z",
      "hook_event_name": "beforeSubmitPrompt",
      "prompt": "Implement token-bucket rate limiting on POST /api/orders. Ticket FIN-3678.",
      "conversation_id": "a1b2c3d4-e5f6-7890-abcd-ef1234567890",
      "cwd": "/Users/dev/gh-org/orders",
      "workspace_roots": ["/Users/dev/gh-org/orders"]
    },
    {
      "eventId": "evt_01HZX9K2M3N4P5Q6R7S8TB",
      "event": "agentThought",
      "ts": "2026-04-08T08:51:06.400Z",
      "hook_event_name": "agentThought",
      "text": "FIN-3678 asks for a token bucket on POST /api/orders. Check the development guidelines, then the orders service, then add the limiter beside the handler.",
      "conversation_id": "a1b2c3d4-e5f6-7890-abcd-ef1234567890"
    },
    {
      "eventId": "evt_01HZX9K2M3N4P5Q6R7S8TA",
      "event": "preToolUse",
      "ts": "2026-04-08T08:51:10.220Z",
      "hook_event_name": "preToolUse",
      "tool_name": "Read",
      "tool_use_id": "toolu_01READGUIDE",
      "tool_input": {
        "path": "/Users/dev/gh-org/orders/docs/development-guidelines.md"
      },
      "conversation_id": "a1b2c3d4-e5f6-7890-abcd-ef1234567890"
    },
    {
      "eventId": "evt_01HZX9K2M3N4P5Q6R7S8TC",
      "event": "preToolUse",
      "ts": "2026-04-08T08:51:14.010Z",
      "hook_event_name": "preToolUse",
      "tool_name": "mcp:orders/list_routes",
      "tool_use_id": "toolu_01MCPROUTES",
      "tool_input": {
        "server": "orders",
        "tool": "list_routes",
        "arguments": { "service": "orders" }
      },
      "conversation_id": "a1b2c3d4-e5f6-7890-abcd-ef1234567890"
    },
    {
      "eventId": "evt_01HZX9K2M3N4P5Q6R7S8TD",
      "event": "postToolUse",
      "ts": "2026-04-08T08:51:14.640Z",
      "hook_event_name": "postToolUse",
      "tool_name": "mcp:orders/list_routes",
      "tool_use_id": "toolu_01MCPROUTES",
      "tool_result": {
        "routes": ["POST /api/orders"]
      },
      "conversation_id": "a1b2c3d4-e5f6-7890-abcd-ef1234567890"
    },
    {
      "eventId": "evt_01HZX9K2M3N4P5Q6R7S8T0",
      "event": "preToolUse",
      "ts": "2026-04-08T08:51:18.441Z",
      "hook_event_name": "preToolUse",
      "tool_name": "Write",
      "tool_use_id": "toolu_01ABCWRITE",
      "tool_input": {
        "path": "/Users/dev/gh-org/orders/src/orders/rate_limit.py",
        "contents": "..."
      },
      "path_hashes": {
        "/Users/dev/gh-org/orders/src/orders/rate_limit.py": null
      },
      "conversation_id": "a1b2c3d4-e5f6-7890-abcd-ef1234567890"
    },
    {
      "eventId": "evt_01HZX9K2M3N4P5Q6R7S8T1",
      "event": "postToolUse",
      "ts": "2026-04-08T08:51:19.102Z",
      "hook_event_name": "postToolUse",
      "tool_name": "Write",
      "tool_use_id": "toolu_01ABCWRITE",
      "path_hashes": {
        "/Users/dev/gh-org/orders/src/orders/rate_limit.py": "e3b0c44298fc1c149afbf4c8996fb92427ae41e4649b934ca495991b7852b855"
      },
      "conversation_id": "a1b2c3d4-e5f6-7890-abcd-ef1234567890"
    },
    {
      "eventId": "evt_01HZX9K2M3N4P5Q6R7S8T2",
      "event": "afterAgentResponse",
      "ts": "2026-04-08T08:52:01.880Z",
      "hook_event_name": "afterAgentResponse",
      "text": "Added token-bucket rate limiting on POST /api/orders in src/orders/rate_limit.py.",
      "conversation_id": "a1b2c3d4-e5f6-7890-abcd-ef1234567890"
    },
    {
      "eventId": "evt_01HZX9K2M3N4P5Q6R7S8T3",
      "event": "sessionEnd",
      "ts": "2026-04-08T08:52:05.010Z"
    }
  ]
}
\end{lstlisting}

\subsubsection{Process evidence}
\label{ex:hitl-evidence}
\begin{lstlisting}[style=jsonex]
{
  "_type": "https://in-toto.io/Statement/v1",
  "subject": [
    {
      "uri": "https://github.com/MYORG/orders/commit/bf02510c8ce0b804a099797510afc19325acfb979538c5b521304c83cde63892",
      "digest": {
        "gitCommit": "bf02510c8ce0b804a099797510afc19325acfb979538c5b521304c83cde63892"
      }
    }
  ],
  "predicateType": "https://jfrog.com/evidence/agentic-process/v0.1",
  "predicate": {
    "processType": "code-development",
    "providers": [
      {
        "harness": {
          "name": "cursor",
          "version": "3.5.33"
        },
        "agent": {
          "id": "stable-agent-id",
          "name": "cursor-agent",
          "version": "1.0"
        },
        "languageModels": [
          {
            "inferenceProvider": "anthropic/claude-opus-4-8",
            "resolved": "anthropic/claude-opus-4-8-20260701"
          }
        ]
      }
    ],
    "traceId": "feature/fin-3678-rate-limit-bf02510",
    "sessionsLogs": [
      {
        "uri": "https://myorg.jfrog.io/artifactory/agentic-session-logs/a1b2c3d4-e5f6-7890-abcd-ef1234567890.json",
        "digest": {
          "sha256": "7c3a1f8e2d94b0c6a15e89d3f4b2a7c0e91d6b48a3f5c2e7d1b9a0c4e6f8d2a1"
        }
      },
      {
        "uri": "https://myorg.jfrog.io/artifactory/agentic-session-logs/b7c1e2a0-1111-4a2b-9c3d-555566667777.json",
        "digest": {
          "sha256": "11ab44c0e8d21f6a9b3c70e5d4a18f2c6b90e7d3a5c1f8e4b2d6a0c9e3f7b851"
        }
      },
      {
        "uri": "https://myorg.jfrog.io/artifactory/agentic-session-logs/c8d2f3b1-2222-4b3c-8d4e-888899990000.json",
        "digest": {
          "sha256": "77e0a91c3b5d6f2e8a4c0d1b7e9f3a6c2d8b4e0f1a5c7d3b9e6f2a8c4d0b1e73"
        }
      }
    ],
    "tools": [
      { "name": "Read", "version": "cursor-native" },
      { "name": "Write", "version": "cursor-native" },
      { "name": "mcp:orders/list_routes", "version": "orders-mcp-1.4.0" }
    ],
    "contextArtifacts": [
      {
        "tags": ["policy"],
        "uri": "https://myorg.jfrog.io/artifactory/policies/development-guidelines.md",
        "digest": {
          "sha256": "3b1f9c2a7e4d80b6c5a192f8e0d47b3c1a6e5f90d2c8b4a7e3f1c0d9b8a75621"
        }
      },
      {
        "tags": ["guideline"],
        "uri": "https://myorg.jfrog.io/artifactory/policies/architecture-guidelines.md",
        "digest": {
          "sha256": "9e2a4c1b8f6d03a7e5c192b0d4f8a6c3e1b7d590a2f4c8e6b0d3a1f7c5e49280"
        }
      }
    ],
    "custom": {
      "baseCommit": {
        "uri": "https://github.com/MYORG/orders/commit/1b94ebd1ef32a4be4480dc70ed8ec2c6d55a2a0c",
        "digest": {
          "gitCommit": "1b94ebd1ef32a4be4480dc70ed8ec2c6d55a2a0c"
        }
      },
      "requirements": [
        {
          "issue": "https://myorg.atlassian.net/browse/FIN-3678",
          "title": "Rate limit POST /api/orders"
        }
      ]
    },
    "result": "COMPLETED",
    "intents": [
      "Implement token-bucket rate limiting on POST /api/orders"
    ],
    "processSummary": "Three sessions on feature branch fin-3678-rate-limit produced commit bf02510. The signed statement cites each session log by digest. Developer and Cursor agent implemented rate limiting and committed for pre-merge review.",
    "owner": "ada@myorg.com",
    "reviewers": ["ada@myorg.com"],
    "startTimestamp": "2026-04-08T08:51:02.114Z",
    "endTimestamp": "2026-04-08T09:10:44.000Z"
  },
  "createdAt": "2026-04-08T09:11:02.000Z",
  "createdBy": "agent-runtime-tool"
}
\end{lstlisting}

\subsubsection{Alignment evidence}
\label{ex:hitl-alignment}
\begin{lstlisting}[style=jsonex]
{
  "_type": "https://in-toto.io/Statement/v1",
  "subject": [
    {
      "uri": "https://github.com/MYORG/orders/commit/bf02510c8ce0b804a099797510afc19325acfb979538c5b521304c83cde63892",
      "digest": {
        "gitCommit": "bf02510c8ce0b804a099797510afc19325acfb979538c5b521304c83cde63892"
      }
    }
  ],
  "predicateType": "https://jfrog.com/evidence/agentic-process/v0.1",
  "predicate": {
    "processType": "session-alignment",
    "providers": [
      {
        "harness": {
          "name": "cursor",
          "version": "3.5.33"
        },
        "agent": {
          "id": "alignment-checker",
          "name": "alignment-checker",
          "version": "1.2.0"
        },
        "languageModels": [
          {
            "inferenceProvider": "anthropic/claude-sonnet-4",
            "resolved": "anthropic/claude-sonnet-4-20250514"
          }
        ]
      }
    ],
    "traceId": "feature/fin-3678-rate-limit-bf02510",
    "sessionsLogs": [
      {
        "uri": "https://myorg.jfrog.io/artifactory/agentic-session-logs/a1b2c3d4-e5f6-7890-abcd-ef1234567890.json",
        "digest": {
          "sha256": "7c3a1f8e2d94b0c6a15e89d3f4b2a7c0e91d6b48a3f5c2e7d1b9a0c4e6f8d2a1"
        }
      },
      {
        "uri": "https://myorg.jfrog.io/artifactory/agentic-session-logs/b7c1e2a0-1111-4a2b-9c3d-555566667777.json",
        "digest": {
          "sha256": "11ab44c0e8d21f6a9b3c70e5d4a18f2c6b90e7d3a5c1f8e4b2d6a0c9e3f7b851"
        }
      },
      {
        "uri": "https://myorg.jfrog.io/artifactory/agentic-session-logs/c8d2f3b1-2222-4b3c-8d4e-888899990000.json",
        "digest": {
          "sha256": "77e0a91c3b5d6f2e8a4c0d1b7e9f3a6c2d8b4e0f1a5c7d3b9e6f2a8c4d0b1e73"
        }
      }
    ],
    "contextArtifacts": [
      {
        "tags": ["policy"],
        "uri": "https://myorg.jfrog.io/artifactory/policies/mandatory-context-policy.md",
        "digest": {
          "sha256": "41c0e93b6d287fa5029b8c1d64a7e3f1c0d9b8a7562104c8e6b0d3a1f7c5e4928"
        }
      },
      {
        "tags": ["policy"],
        "uri": "https://myorg.jfrog.io/artifactory/policies/development-guidelines.md",
        "digest": {
          "sha256": "3b1f9c2a7e4d80b6c5a192f8e0d47b3c1a6e5f90d2c8b4a7e3f1c0d9b8a75621"
        }
      },
      {
        "tags": ["guideline"],
        "uri": "https://myorg.jfrog.io/artifactory/policies/architecture-guidelines.md",
        "digest": {
          "sha256": "9e2a4c1b8f6d03a7e5c192b0d4f8a6c3e1b7d590a2f4c8e6b0d3a1f7c5e49280"
        }
      }
    ],
    "custom": {
      "baseCommit": {
        "uri": "https://github.com/MYORG/orders/commit/1b94ebd1ef32a4be4480dc70ed8ec2c6d55a2a0c",
        "digest": {
          "gitCommit": "1b94ebd1ef32a4be4480dc70ed8ec2c6d55a2a0c"
        }
      },
      "policyResults": [
        {
          "policy": "https://myorg.jfrog.io/artifactory/policies/mandatory-context-policy.md",
          "result": "ALIGNED"
        },
        {
          "policy": "https://myorg.jfrog.io/artifactory/policies/development-guidelines.md",
          "result": "ALIGNED"
        },
        {
          "policy": "https://myorg.jfrog.io/artifactory/policies/architecture-guidelines.md",
          "result": "ALIGNED"
        }
      ]
    },
    "result": "ALIGNED",
    "owner": "platform-agent-ops@myorg.com",
    "startTimestamp": "2026-04-08T09:10:50.000Z",
    "endTimestamp": "2026-04-08T09:11:05.000Z"
  },
  "createdAt": "2026-04-08T09:11:10.000Z",
  "createdBy": "alignment-checker"
}
\end{lstlisting}

\subsection{Autonomous code development}

The autonomous example uses the same object types. Its process evidence has
an owner and no human reviewer; its alignment evidence is scoped to one
session log.

\subsubsection{Session log}
\begin{lstlisting}[style=jsonex]
{
  "sessionId": "a1b2c3d4-e5f6-7890-abcd-ef1234567890",
  "affectedEntities": [
    {
      "gitCommit": "bf02510c8ce0b804a099797510afc19325acfb979538c5b521304c83cde63892"
    }
  ],
  "provider": {
    "harness": {
      "name": "cursor",
      "version": "3.5.33"
    },
    "agent": {
      "id": "stable-agent-id",
      "name": "cursor-agent",
      "version": "1.0"
    },
    "languageModels": [
      {
        "inferenceProvider": "anthropic/claude-opus-4-8",
        "resolved": "anthropic/claude-opus-4-8-20260701"
      }
    ]
  },
  "timeline": [
    {
      "eventId": "evt_01HZX9K2M3N4P5Q6R7S8T9",
      "event": "beforeSubmitPrompt",
      "ts": "2026-04-08T08:51:02.114Z",
      "hook_event_name": "beforeSubmitPrompt",
      "prompt": "Implement token-bucket rate limiting on POST /api/orders. Ticket FIN-3678.",
      "conversation_id": "a1b2c3d4-e5f6-7890-abcd-ef1234567890",
      "cwd": "/Users/dev/gh-org/orders",
      "workspace_roots": ["/Users/dev/gh-org/orders"]
    },
    {
      "eventId": "evt_01HZX9K2M3N4P5Q6R7S8TA",
      "event": "preToolUse",
      "ts": "2026-04-08T08:51:10.220Z",
      "hook_event_name": "preToolUse",
      "tool_name": "Read",
      "tool_use_id": "toolu_01READGUIDE",
      "tool_input": {
        "path": "/Users/dev/gh-org/orders/docs/development-guidelines.md"
      },
      "conversation_id": "a1b2c3d4-e5f6-7890-abcd-ef1234567890"
    },
    {
      "eventId": "evt_01HZX9K2M3N4P5Q6R7S8T0",
      "event": "preToolUse",
      "ts": "2026-04-08T08:51:18.441Z",
      "hook_event_name": "preToolUse",
      "tool_name": "Write",
      "tool_use_id": "toolu_01ABCWRITE",
      "tool_input": {
        "path": "/Users/dev/gh-org/orders/src/orders/rate_limit.py",
        "contents": "..."
      },
      "path_hashes": {
        "/Users/dev/gh-org/orders/src/orders/rate_limit.py": null
      },
      "conversation_id": "a1b2c3d4-e5f6-7890-abcd-ef1234567890"
    },
    {
      "eventId": "evt_01HZX9K2M3N4P5Q6R7S8T1",
      "event": "postToolUse",
      "ts": "2026-04-08T08:51:19.102Z",
      "hook_event_name": "postToolUse",
      "tool_name": "Write",
      "tool_use_id": "toolu_01ABCWRITE",
      "path_hashes": {
        "/Users/dev/gh-org/orders/src/orders/rate_limit.py": "e3b0c44298fc1c149afbf4c8996fb92427ae41e4649b934ca495991b7852b855"
      },
      "conversation_id": "a1b2c3d4-e5f6-7890-abcd-ef1234567890"
    },
    {
      "eventId": "evt_01HZX9K2M3N4P5Q6R7S8T2",
      "event": "afterAgentResponse",
      "ts": "2026-04-08T08:52:01.880Z",
      "hook_event_name": "afterAgentResponse",
      "text": "Added token-bucket rate limiting on POST /api/orders in src/orders/rate_limit.py.",
      "conversation_id": "a1b2c3d4-e5f6-7890-abcd-ef1234567890"
    },
    {
      "eventId": "evt_01HZX9K2M3N4P5Q6R7S8T3",
      "event": "sessionEnd",
      "ts": "2026-04-08T08:52:05.010Z"
    }
  ]
}
\end{lstlisting}

\subsubsection{Process evidence}
\label{ex:auto-evidence}
\begin{lstlisting}[style=jsonex]
{
  "_type": "https://in-toto.io/Statement/v1",
  "subject": [
    {
      "uri": "https://github.com/MYORG/orders/commit/bf02510c8ce0b804a099797510afc19325acfb979538c5b521304c83cde63892",
      "digest": {
        "gitCommit": "bf02510c8ce0b804a099797510afc19325acfb979538c5b521304c83cde63892"
      }
    }
  ],
  "predicateType": "https://jfrog.com/evidence/agentic-process/v0.1",
  "predicate": {
    "processType": "code-development",
    "providers": [
      {
        "harness": {
          "name": "cursor",
          "version": "3.5.33"
        },
        "agent": {
          "id": "stable-agent-id",
          "name": "cursor-agent",
          "version": "1.0"
        },
        "languageModels": [
          {
            "inferenceProvider": "anthropic/claude-opus-4-8",
            "resolved": "anthropic/claude-opus-4-8-20260701"
          }
        ]
      }
    ],
    "traceId": "feature/fin-3678-rate-limit-bf02510",
    "sessionsLogs": [
      {
        "uri": "https://myorg.jfrog.io/artifactory/agentic-session-logs/a1b2c3d4-e5f6-7890-abcd-ef1234567890.json",
        "digest": {
          "sha256": "7c3a1f8e2d94b0c6a15e89d3f4b2a7c0e91d6b48a3f5c2e7d1b9a0c4e6f8d2a1"
        }
      }
    ],
    "tools": [
      { "name": "Read", "version": "cursor-native" },
      { "name": "Write", "version": "cursor-native" }
    ],
    "contextArtifacts": [
      {
        "tags": ["policy"],
        "uri": "https://myorg.jfrog.io/artifactory/policies/development-guidelines.md",
        "digest": {
          "sha256": "3b1f9c2a7e4d80b6c5a192f8e0d47b3c1a6e5f90d2c8b4a7e3f1c0d9b8a75621"
        }
      },
      {
        "tags": ["guideline"],
        "uri": "https://myorg.jfrog.io/artifactory/policies/architecture-guidelines.md",
        "digest": {
          "sha256": "9e2a4c1b8f6d03a7e5c192b0d4f8a6c3e1b7d590a2f4c8e6b0d3a1f7c5e49280"
        }
      }
    ],
    "custom": {
      "baseCommit": {
        "uri": "https://github.com/MYORG/orders/commit/1b94ebd1ef32a4be4480dc70ed8ec2c6d55a2a0c",
        "digest": {
          "gitCommit": "1b94ebd1ef32a4be4480dc70ed8ec2c6d55a2a0c"
        }
      },
      "requirements": [
        {
          "issue": "https://myorg.atlassian.net/browse/FIN-3678",
          "title": "Rate limit POST /api/orders"
        }
      ]
    },
    "result": "COMPLETED",
    "intents": [
      "Implement token-bucket rate limiting on POST /api/orders"
    ],
    "processSummary": "One agent session on feature branch fin-3678-rate-limit produced commit bf02510. The agent implemented rate limiting and committed it for pre-merge review. The statement names an owner and no human reviewer.",
    "owner": "ada@myorg.com",
    "startTimestamp": "2026-04-08T08:51:02.114Z",
    "endTimestamp": "2026-04-08T09:10:44.000Z"
  },
  "createdAt": "2026-04-08T09:11:02.000Z",
  "createdBy": "agent-runtime-tool"
}
\end{lstlisting}

\subsubsection{Alignment evidence}
\begin{lstlisting}[style=jsonex]
{
  "_type": "https://in-toto.io/Statement/v1",
  "subject": [
    {
      "uri": "https://myorg.jfrog.io/artifactory/agentic-session-logs/a1b2c3d4-e5f6-7890-abcd-ef1234567890.json",
      "digest": {
        "sha256": "7c3a1f8e2d94b0c6a15e89d3f4b2a7c0e91d6b48a3f5c2e7d1b9a0c4e6f8d2a1"
      }
    }
  ],
  "predicateType": "https://jfrog.com/evidence/agentic-process/v0.1",
  "predicate": {
    "processType": "session-alignment",
    "providers": [
      {
        "harness": {
          "name": "cursor",
          "version": "3.5.33"
        },
        "agent": {
          "id": "alignment-checker",
          "name": "alignment-checker",
          "version": "1.2.0"
        },
        "languageModels": [
          {
            "inferenceProvider": "anthropic/claude-sonnet-4",
            "resolved": "anthropic/claude-sonnet-4-20250514"
          }
        ]
      }
    ],
    "contextArtifacts": [
      {
        "tags": ["policy"],
        "uri": "https://myorg.jfrog.io/artifactory/policies/mandatory-context-policy.md",
        "digest": {
          "sha256": "41c0e93b6d287fa5029b8c1d64a7e3f1c0d9b8a7562104c8e6b0d3a1f7c5e4928"
        }
      },
      {
        "tags": ["policy"],
        "uri": "https://myorg.jfrog.io/artifactory/policies/development-guidelines.md",
        "digest": {
          "sha256": "3b1f9c2a7e4d80b6c5a192f8e0d47b3c1a6e5f90d2c8b4a7e3f1c0d9b8a75621"
        }
      },
      {
        "tags": ["guideline"],
        "uri": "https://myorg.jfrog.io/artifactory/policies/architecture-guidelines.md",
        "digest": {
          "sha256": "9e2a4c1b8f6d03a7e5c192b0d4f8a6c3e1b7d590a2f4c8e6b0d3a1f7c5e49280"
        }
      }
    ],
    "custom": {
      "baseCommit": {
        "uri": "https://github.com/MYORG/orders/commit/1b94ebd1ef32a4be4480dc70ed8ec2c6d55a2a0c",
        "digest": {
          "gitCommit": "1b94ebd1ef32a4be4480dc70ed8ec2c6d55a2a0c"
        }
      }
    },
    "result": "ALIGNED",
    "owner": "platform-agent-ops@myorg.com",
    "startTimestamp": "2026-04-08T09:10:50.000Z",
    "endTimestamp": "2026-04-08T09:11:05.000Z"
  },
  "createdAt": "2026-04-08T09:11:10.000Z",
  "createdBy": "alignment-checker"
}
\end{lstlisting}

\subsection{Code review}

\subsubsection{Session log}
\begin{lstlisting}[style=jsonex]
{
  "sessionId": "b2c3d4e5-f6a7-8901-bcde-f12345678901",
  "affectedEntities": [
    {
      "gitCommit": "bf02510c8ce0b804a099797510afc19325acfb979538c5b521304c83cde63892"
    }
  ],
  "provider": {
    "harness": {
      "name": "github-actions",
      "version": "2.322.0"
    },
    "agent": {
      "id": "code-review-agent",
      "name": "code-review-agent",
      "version": "0.9.0"
    },
    "languageModels": [
      {
        "inferenceProvider": "anthropic/claude-sonnet-4",
        "resolved": "anthropic/claude-sonnet-4-20250514"
      }
    ]
  },
  "timeline": [
    {
      "eventId": "evt_rev_01",
      "event": "beforeSubmitPrompt",
      "ts": "2026-04-08T09:10:00.000Z",
      "hook_event_name": "beforeSubmitPrompt",
      "prompt": "Review commit bf02510 against org review rubric. Classify change risk.",
      "run_id": "b2c3d4e5-f6a7-8901-bcde-f12345678901"
    },
    {
      "eventId": "evt_rev_02",
      "event": "preToolUse",
      "ts": "2026-04-08T09:10:05.220Z",
      "hook_event_name": "preToolUse",
      "tool_name": "Read",
      "tool_use_id": "toolu_rev_rubric",
      "tool_input": {
        "path": "org-review-rubric.md"
      },
      "run_id": "b2c3d4e5-f6a7-8901-bcde-f12345678901"
    },
    {
      "eventId": "evt_rev_03",
      "event": "preToolUse",
      "ts": "2026-04-08T09:10:18.440Z",
      "hook_event_name": "preToolUse",
      "tool_name": "GitDiff",
      "tool_use_id": "toolu_rev_diff",
      "tool_input": {
        "baseCommit": "1b94ebd1ef32a4be4480dc70ed8ec2c6d55a2a0c",
        "headCommit": "bf02510c8ce0b804a099797510afc19325acfb979538c5b521304c83cde63892"
      },
      "run_id": "b2c3d4e5-f6a7-8901-bcde-f12345678901"
    },
    {
      "eventId": "evt_rev_04",
      "event": "afterAgentResponse",
      "ts": "2026-04-08T09:12:35.000Z",
      "hook_event_name": "afterAgentResponse",
      "text": "Change adds token-bucket rate limiting on POST /api/orders in a new file, src/orders/rate_limit.py, as FIN-3678 requires. Classification: STANDARD. Approved. Not eligible for auto-merge; merge requires human approval.",
      "run_id": "b2c3d4e5-f6a7-8901-bcde-f12345678901"
    },
    {
      "eventId": "evt_rev_05",
      "event": "sessionEnd",
      "ts": "2026-04-08T09:12:41.000Z"
    }
  ]
}
\end{lstlisting}

\subsubsection{Process evidence}
\label{ex:review-evidence}
\begin{lstlisting}[style=jsonex]
{
  "_type": "https://in-toto.io/Statement/v1",
  "subject": [
    {
      "uri": "https://github.com/MYORG/orders/commit/bf02510c8ce0b804a099797510afc19325acfb979538c5b521304c83cde63892",
      "digest": {
        "gitCommit": "bf02510c8ce0b804a099797510afc19325acfb979538c5b521304c83cde63892"
      }
    }
  ],
  "predicateType": "https://jfrog.com/evidence/agentic-process/v0.1",
  "predicate": {
    "processType": "code-review",
    "providers": [
      {
        "harness": {
          "name": "github-actions",
          "version": "2.322.0"
        },
        "agent": {
          "id": "code-review-agent",
          "name": "code-review-agent",
          "version": "0.9.0"
        },
        "languageModels": [
          {
            "inferenceProvider": "anthropic/claude-sonnet-4",
            "resolved": "anthropic/claude-sonnet-4-20250514"
          }
        ]
      }
    ],
    "traceId": "b2c3d4e5-f6a7-8901-bcde-f12345678901",
    "sessionsLogs": [
      {
        "uri": "https://myorg.jfrog.io/artifactory/agentic-session-logs/b2c3d4e5-f6a7-8901-bcde-f12345678901.json",
        "digest": {
          "sha256": "1a9c4e2b7d80f3a6c5e192d8b0a47c3e1f6d5b90c2a8e4b7d3f1c0a9e8b75630"
        }
      }
    ],
    "tools": [
      { "name": "Read", "version": "review-agent" },
      { "name": "GitDiff", "version": "review-agent" }
    ],
    "contextArtifacts": [
      {
        "tags": ["policy"],
        "uri": "https://myorg.jfrog.io/artifactory/policies/org-review-rubric.md",
        "digest": {
          "sha256": "6d2e8a1c4f90b3d7e5a192c0f8b46a3d1c7e5f80b2d4a8c6e0f3b1d9a8c54721"
        }
      }
    ],
    "custom": {
      "baseCommit": {
        "uri": "https://github.com/MYORG/orders/commit/1b94ebd1ef32a4be4480dc70ed8ec2c6d55a2a0c",
        "digest": {
          "gitCommit": "1b94ebd1ef32a4be4480dc70ed8ec2c6d55a2a0c"
        }
      },
      "requirements": [
        {
          "issue": "https://myorg.atlassian.net/browse/FIN-3678",
          "title": "Rate limit POST /api/orders"
        }
      ],
      "changeProfile": {
        "classification": "STANDARD",
        "filesChanged": 1,
        "linesAdded": 42,
        "linesRemoved": 0,
        "autoMergeEligible": false
      },
      "pullRequest": "https://github.com/MYORG/orders/pull/412"
    },
    "result": "APPROVED",
    "intents": [
      "Code review agent classification of commit bf02510"
    ],
    "processSummary": "Code review agent inspected diff 1b94ebd1..bf02510, a new token-bucket rate limiter for FIN-3678, classified the change as STANDARD, and approved it. Not eligible for auto-merge; the merge requires human approval.",
    "owner": "platform-agent-ops@myorg.com",
    "startTimestamp": "2026-04-08T09:10:00.000Z",
    "endTimestamp": "2026-04-08T09:12:41.000Z"
  },
  "createdAt": "2026-04-08T09:13:05.000Z",
  "createdBy": "agent-runtime-tool"
}
\end{lstlisting}

\subsubsection{Alignment evidence}
\begin{lstlisting}[style=jsonex]
{
  "_type": "https://in-toto.io/Statement/v1",
  "subject": [
    {
      "uri": "https://myorg.jfrog.io/artifactory/agentic-session-logs/b2c3d4e5-f6a7-8901-bcde-f12345678901.json",
      "digest": {
        "sha256": "1a9c4e2b7d80f3a6c5e192d8b0a47c3e1f6d5b90c2a8e4b7d3f1c0a9e8b75630"
      }
    }
  ],
  "predicateType": "https://jfrog.com/evidence/agentic-process/v0.1",
  "predicate": {
    "processType": "session-alignment",
    "providers": [
      {
        "harness": {
          "name": "github-actions",
          "version": "2.322.0"
        },
        "agent": {
          "id": "alignment-checker",
          "name": "alignment-checker",
          "version": "1.2.0"
        },
        "languageModels": [
          {
            "inferenceProvider": "anthropic/claude-sonnet-4",
            "resolved": "anthropic/claude-sonnet-4-20250514"
          }
        ]
      }
    ],
    "contextArtifacts": [
      {
        "tags": ["policy"],
        "uri": "https://myorg.jfrog.io/artifactory/policies/org-review-rubric.md",
        "digest": {
          "sha256": "6d2e8a1c4f90b3d7e5a192c0f8b46a3d1c7e5f80b2d4a8c6e0f3b1d9a8c54721"
        }
      }
    ],
    "custom": {
      "baseCommit": {
        "uri": "https://github.com/MYORG/orders/commit/1b94ebd1ef32a4be4480dc70ed8ec2c6d55a2a0c",
        "digest": {
          "gitCommit": "1b94ebd1ef32a4be4480dc70ed8ec2c6d55a2a0c"
        }
      }
    },
    "result": "ALIGNED",
    "owner": "platform-agent-ops@myorg.com",
    "startTimestamp": "2026-04-08T09:12:50.000Z",
    "endTimestamp": "2026-04-08T09:13:02.000Z"
  },
  "createdAt": "2026-04-08T09:13:08.000Z",
  "createdBy": "alignment-checker"
}
\end{lstlisting}

\subsection{Customer success}

\subsubsection{Session log}
\begin{lstlisting}[style=jsonex]
{
  "sessionId": "c9e8d7f6-1111-2222-3333-444455556666",
  "affectedEntities": [
    {
      "sessionId": "c9e8d7f6-1111-2222-3333-444455556666"
    }
  ],
  "provider": {
    "harness": {
      "name": "support-desk",
      "version": "4.2.0"
    },
    "agent": {
      "id": "support-agent",
      "name": "support-agent",
      "version": "2.1.0"
    },
    "languageModels": [
      {
        "inferenceProvider": "openai/gpt-4.1",
        "resolved": "openai/gpt-4.1-2025-04-14"
      }
    ]
  },
  "timeline": [
    {
      "eventId": "evt_sup_01",
      "event": "beforeSubmitPrompt",
      "ts": "2026-04-08T14:00:02.000Z",
      "hook_event_name": "beforeSubmitPrompt",
      "prompt": "Can I get a refund on order 44119? It has been 20 days.",
      "ticket": "SUP-8821"
    },
    {
      "eventId": "evt_sup_02",
      "event": "preToolUse",
      "ts": "2026-04-08T14:00:04.200Z",
      "hook_event_name": "preToolUse",
      "tool_name": "Read",
      "tool_use_id": "toolu_sup_policy",
      "tool_input": {
        "path": "support-refund-policy.md"
      }
    },
    {
      "eventId": "evt_sup_03",
      "event": "preToolUse",
      "ts": "2026-04-08T14:00:06.800Z",
      "hook_event_name": "preToolUse",
      "tool_name": "OrderLookup",
      "tool_use_id": "toolu_sup_order",
      "tool_input": {
        "orderId": "44119"
      }
    },
    {
      "eventId": "evt_sup_04",
      "event": "afterAgentResponse",
      "ts": "2026-04-08T14:00:18.000Z",
      "hook_event_name": "afterAgentResponse",
      "text": "Order 44119 is outside the 14-day refund window. A refund cannot be issued under current policy."
    },
    {
      "eventId": "evt_sup_05",
      "event": "sessionEnd",
      "ts": "2026-04-08T14:00:19.000Z"
    }
  ]
}
\end{lstlisting}

\subsubsection{Process evidence}
\label{ex:support-evidence}
\begin{lstlisting}[style=jsonex]
{
  "_type": "https://in-toto.io/Statement/v1",
  "subject": [
    {
      "uri": "https://myorg.jfrog.io/artifactory/agentic-session-logs/c9e8d7f6-1111-2222-3333-444455556666.json",
      "digest": {
        "sha256": "8b4d1e7a2c90f5d3e6a182b9c0d47e3f1a6c5b80d2e8a4c7f3b1d0e9a8c64712"
      }
    }
  ],
  "predicateType": "https://jfrog.com/evidence/agentic-process/v0.1",
  "predicate": {
    "processType": "support-process",
    "providers": [
      {
        "harness": {
          "name": "support-desk",
          "version": "4.2.0"
        },
        "agent": {
          "id": "support-agent",
          "name": "support-agent",
          "version": "2.1.0"
        },
        "languageModels": [
          {
            "inferenceProvider": "openai/gpt-4.1",
            "resolved": "openai/gpt-4.1-2025-04-14"
          }
        ]
      }
    ],
    "traceId": "c9e8d7f6-1111-2222-3333-444455556666",
    "tools": [
      { "name": "Read", "version": "support-agent" },
      { "name": "OrderLookup", "version": "support-agent" }
    ],
    "contextArtifacts": [
      {
        "tags": ["policy"],
        "uri": "https://myorg.jfrog.io/artifactory/policies/support-refund-policy.md",
        "digest": {
          "sha256": "2f7c1a9e4b80d3c6e5a192f0d8b46c3a1e7d5f90b2c4a8e6d0f3b1c9a8e54730"
        }
      }
    ],
    "custom": {
      "ticket": "SUP-8821",
      "orderId": "44119"
    },
    "result": "COMPLETED",
    "intents": [
      "Determine refund eligibility for order 44119"
    ],
    "processSummary": "Autonomous support agent looked up order 44119, applied the 14-day refund policy, and denied the refund. Single session; no SDLC subject.",
    "owner": "support-platform@myorg.com",
    "startTimestamp": "2026-04-08T14:00:02.000Z",
    "endTimestamp": "2026-04-08T14:00:19.000Z"
  },
  "createdAt": "2026-04-08T14:00:25.000Z",
  "createdBy": "agent-runtime-tool"
}
\end{lstlisting}

\subsubsection{Alignment evidence}
\begin{lstlisting}[style=jsonex]
{
  "_type": "https://in-toto.io/Statement/v1",
  "subject": [
    {
      "uri": "https://myorg.jfrog.io/artifactory/agentic-session-logs/c9e8d7f6-1111-2222-3333-444455556666.json",
      "digest": {
        "sha256": "8b4d1e7a2c90f5d3e6a182b9c0d47e3f1a6c5b80d2e8a4c7f3b1d0e9a8c64712"
      }
    }
  ],
  "predicateType": "https://jfrog.com/evidence/agentic-process/v0.1",
  "predicate": {
    "processType": "session-alignment",
    "providers": [
      {
        "harness": {
          "name": "support-desk",
          "version": "4.2.0"
        },
        "agent": {
          "id": "alignment-checker",
          "name": "alignment-checker",
          "version": "1.2.0"
        },
        "languageModels": [
          {
            "inferenceProvider": "anthropic/claude-sonnet-4",
            "resolved": "anthropic/claude-sonnet-4-20250514"
          }
        ]
      }
    ],
    "contextArtifacts": [
      {
        "tags": ["policy"],
        "uri": "https://myorg.jfrog.io/artifactory/policies/support-refund-policy.md",
        "digest": {
          "sha256": "2f7c1a9e4b80d3c6e5a192f0d8b46c3a1e7d5f90b2c4a8e6d0f3b1c9a8e54730"
        }
      }
    ],
    "custom": {
      "ticket": "SUP-8821",
      "orderId": "44119"
    },
    "result": "ALIGNED",
    "owner": "support-platform@myorg.com",
    "startTimestamp": "2026-04-08T14:00:26.000Z",
    "endTimestamp": "2026-04-08T14:00:34.000Z"
  },
  "createdAt": "2026-04-08T14:00:36.000Z",
  "createdBy": "alignment-checker"
}
\end{lstlisting}

\end{document}